\documentclass[conference]{IEEEtran}
\IEEEoverridecommandlockouts
\usepackage{cite}
\usepackage{amsmath,amssymb,amsfonts}
\usepackage{algorithmic}
\usepackage{graphicx}
\usepackage{textcomp}
\usepackage{xcolor}
\usepackage[hidelinks]{hyperref}
\usepackage{tikz}
\usetikzlibrary{positioning,shapes,arrows}
\def\BibTeX{{\rm B\kern-.05em{\sc i\kern-.025em b}\kern-.08em
    T\kern-.1667em\lower.7ex\hbox{E}\kern-.125emX}}
\begin{document}

\title{TempDiffReg: Temporal Diffusion Model for Non-Rigid 2D-3D Vascular Registration

\thanks{
This work was supported in part by the Research Grants Council of the Hong Kong Special Administrative Region, China (Grant No. T45-401/22-N), in part by Shenzhen Science and Technology Program (Grant No. RCYX20231211090127030), in part by National Natural Science Foundation of China (Grant Nos. 62372441, U22A2034 and 62176016, 72274127), in part by Guangdong Basic and Applied Basic Research Foundation (Grant No. 2023A1515030268), in part by National Key R\&D Program of China (Grant No. 2021YFB2104800), in part by Guizhou Province Science and Technology Project (Grant No. Qiankehe [2024] General 058), in part by Capital Health Development Research Project (Grant No. 2022-2-2013), and in part by Haidian innovation and translation program (Grant No. HDCXZHKC2023203).
}
}
\author{
Zehua Liu$^{1, 2}$\dag,
Shihao Zou$^{3}$\dag,
Jincai Huang$^{4}$,
Yanfang Zhang$^{5}$,
\\
Chao Tong$^{1, 2}$,
Weixin Si$^{4}$*, \IEEEmembership{Senior Member, IEEE}
\\
\IEEEauthorblockA{$^{1}$School of Computer Science and Engineering, Beihang University, Beijing, China \\}
\IEEEauthorblockA{$^{2}$State Key Laboratory of Virtual Reality Technology and Systems, Beihang University, Beijing, China \\}
\IEEEauthorblockA{$^{3}$Shenzhen Institutes of Advanced Technology, Chinese Academy of Sciences, Shenzhen, China \\}
\IEEEauthorblockA{$^{4}$School of Computer Science and Control Engineering, Shenzhen University of Advanced Technology, Shenzhen, China \\}
\IEEEauthorblockA{$^{5}$Department of Interventional Radiology, Shenzhen People’s Hospital, Shenzhen, China \\}

\IEEEauthorblockA{$\dag$ These authors contributed equally. \quad $*$ Corresponding author: siweixin@suat-sz.edu.cn \\}

}

\maketitle

\begin{abstract}
Transarterial chemoembolization (TACE) is a preferred treatment option for hepatocellular carcinoma and other liver malignancies, yet it remains a highly challenging procedure due to complex intra-operative vascular navigation and anatomical variability.
Accurate and robust 2D-3D vessel registration is essential to guide microcatheter and instruments during TACE, enabling precise localization of vascular structures and optimal therapeutic targeting.
To tackle this issue, we develop a coarse-to-fine registration strategy. First, we introduce a global alignment module, structure-aware perspective n-point (SA-PnP), to establish correspondence between 2D and 3D vessel structures. Second, we propose TempDiffReg, a temporal diffusion model that performs vessel deformation iteratively by leveraging temporal context to capture complex anatomical variations and local structural changes.
We collected data from 23 patients and constructed 626 paired multi-frame samples for comprehensive evaluation.
Experimental results demonstrate that the proposed method consistently outperforms state-of-the-art (SOTA) methods in both accuracy and anatomical plausibility. Specifically, our method achieves a mean squared error (MSE) of 0.63 mm and a mean absolute error (MAE) of 0.51 mm in registration accuracy, representing 66.7\% lower MSE and 17.7\% lower MAE compared to the most competitive existing approaches.
It has the potential to assist less-experienced clinicians in safely and efficiently performing complex TACE procedures, ultimately enhancing both surgical outcomes and patient care. Code and data are available at: \textcolor{blue}{\url{https://github.com/LZH970328/TempDiffReg.git}}
\end{abstract}

\section{Introduction}
Transarterial chemoembolization (TACE) is a widely used interventional procedure for the treatment of liver malignancies~\cite{patel2024locoregional,magyar2025precision}. During TACE, the operator is required to advance a microcatheter into the tumor-supplying arterial branch and deliver chemotherapeutic agents to achieve targeted embolization. Therefore, precise localization and navigation of blood vessels are crucial for the efficient and safe completion of the procedure~\cite{sun2024multi}. However, the absence of accurate three-dimensional (3D) visualization of vascular structures makes it difficult for interventional radiologists to accurately identify and super-select the appropriate branches~\cite{schwab2024portal}. This limitation often necessitates repeated contrast injections and prolongs the procedure time, thereby increasing the exposure of both patients and clinicians to ionizing radiation and elevating the risk of procedure-related complications, such as vessel wall irritation or inflammation~\cite{kim2024otmorph}.

Over the years, numerous approaches have been proposed to tackle non-rigid image registration. Vercauteren et al.~\cite{vercauteren2009diffeomorphic} introduced a non-parametric diffeomorphic image registration algorithm, enabling flexible non-rigid deformation with a mathematical foundation. Zhang et al.~\cite{zhang2023enhancing} developed a multi-resolution fused regular step gradient descent (MR-RSGD) optimization strategy to improve registration performance. Traditionally, image registration has been accomplished by iteratively solving optimization problems, resulting in methods that are theoretically well-founded and effective in many cases, especially for rigid alignment or when good initialization is available~\cite{tang2020icp,Lv2023}. Nevertheless, when the initial alignment is poor, they are more likely to suffer from non-convex optimization issues, become trapped in local minima, and fail to achieve globally optimal solutions~\cite{chen2025survey}.

In the past decade, deep learning methods have significantly improved the accuracy and efficiency of image registration by training general networks with global objective functions on large datasets, in contrast to traditional approaches~\cite{chen2025survey,chen2022transmorph}.
For example, Tian et al.~\cite{tian2024unigradicon} proposed uniGradICON, a foundation model for medical image registration that unifies the speed and accuracy of deep learning with the broad applicability of traditional methods.
Chen et al.~\cite{vitvnet} combines Vision Transformers and convolutional networks to leverage long-range context while preserving spatial detail, and has shown competitive performance in medical image registration.
SIRU-Net~\cite{SIRU} integrates inception and residual modules with Gaussian smoothing to enhance registration accuracy under limited data.

However, the complex branching of hepatic vessels, significant anatomical variability, and substantial discrepancies between pre-operative and intra-operative imaging modalities pose major challenges to accurate vascular structure matching. Consequently, there remains an urgent need for a robust and reliable registration framework in clinical practice~\cite{chen2025survey}.

\begin{figure*}[htb!]%
\centering
\includegraphics[width=0.9\textwidth]{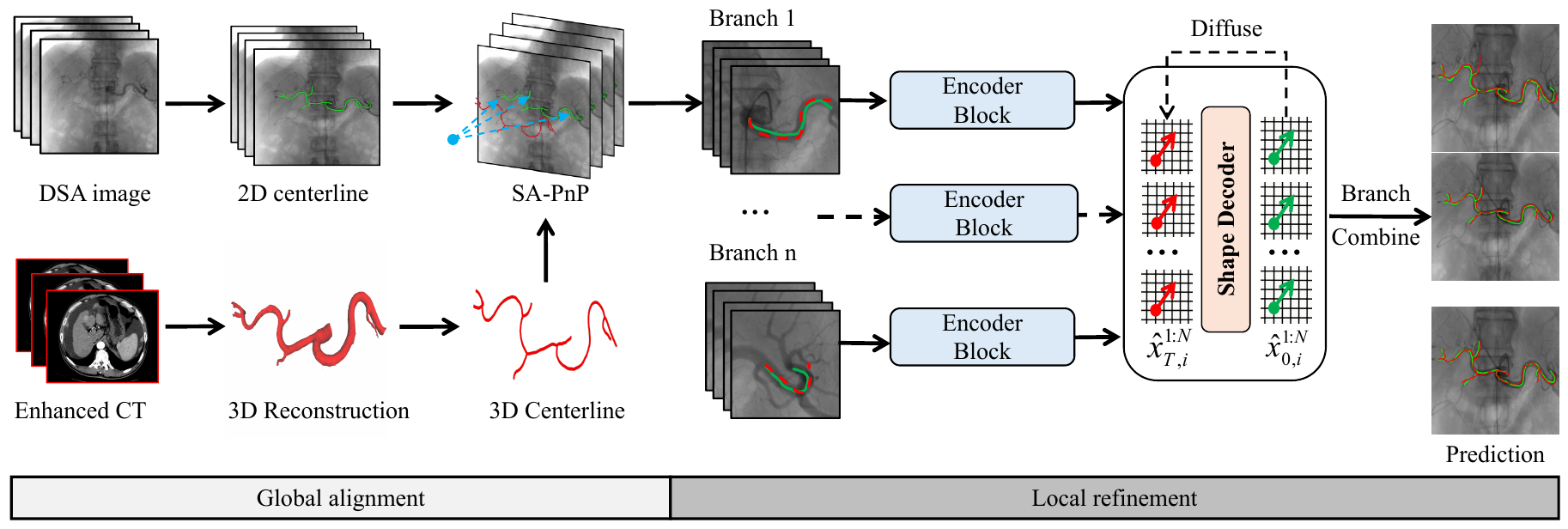}
\caption{
Overview of the proposed coarse-to-fine 2D–3D vessel registration framework. 3D centerlines are reconstructed from pre-operative CTA, while 2D centerlines are manually annotated on intra-operative DSA frames. In the global alignment phase, the SA-PnP module computes an initial pose for each frame. Subsequently, in the local refinement stage, TempDiffReg applies branch-wise non-rigid deformation using temporal diffusion modeling. The refined branches are combined to generate the final registered vessel structure aligned to the target DSA frame.}

\label{fig:Framework}
\end{figure*}

Recently, diffusion-based generative models have emerged as a highly effective framework for probabilistic data modeling, and have been widely applied in class- and text-conditioned image generation~\cite{ho2020denoising}, super-resolution, inpainting~\cite{Rombach_2022_CVPR}, and 3D object generation~\cite{saito2019pifu,Rombach_2022_CVPR}. Recent studies have shown promise in structured data domains such as 3D human modeling~\cite{Corona_2021_CVPR,chen2022gdna,zou2025generating} and iterative stereo correspondence~\cite{shao2022diffustereo}, demonstrating their ability to model intricate geometric transformations. These models operate by gradually transforming a simple noise distribution, such as Gaussian noise, into complex data distributions through a sequence of learned denoising steps. This iterative denoising process enables the synthesis of highly realistic outputs and the recovery of fine structural details, even in tasks involving severe missing information or ambiguity. Despite their remarkable potential, most existing diffusion-based methods are mainly limited to static 2D or 3D settings, or unconditional generation, and often struggle with modeling temporally coherent, structured transformations required for medical registration. 

To address the aforementioned challenges, we propose a coarse-to-fine registration pipeline. Our main contributions are as follows.

\textbf{SA-PnP-based Dataset Generation}: We develop a structure-aware PnP (SA-PnP) algorithm that leverages anatomical features for robust global alignment between 2D and 3D vessels. This facilitates the construction of a dedicated registration dataset with 626 multi-frame samples from 23 clinical cases.

\textbf{TempDiffReg for Local Refinement}: We propose TempDiffReg, a temporal diffusion model that performs branch-wise deformation across frames. Temporal conditioning enhances motion continuity and resolves occlusions and ambiguities in vessel shape.

Experimental results demonstrate that our method consistently outperforms state-of-the-art baselines across five evaluation metrics, achieving 0.63 mm mean squared error (MSE) and 0.51 mm mean absolute error (MAE), which represent 66.7\% and 17.7\% reductions over the best baseline, respectively.

\section{Method}
As shown in Fig.~\ref{fig:Framework}, our method adopts a coarse-to-fine registration pipeline.
In the global alignment stage, we first manually annotate 2D vessel centerlines on intra-operative digital subtraction angiography (DSA) images, and reconstruct the 3D vascular centerlines from arterial-phase contrast-enhanced computed tomography angiography (CTA) using the Vascular Modeling Toolkit (VMTK)~\cite{tian2025bi}. 
Considering that traditional registration methods perform efficiently with good initialization~\cite{song2024iterative}, we develop a structure-aware PnP (SA-PnP) module to generate multi-frame samples with reliable pose parameters.
Next, in the local refinement stage, the vascular tree is decomposed into anatomically meaningful branches, and a temporal diffusion model is employed to capture local deformations for each branch. The outputs from all branches are subsequently combined to reconstruct the final, registered vessel structure in the target DSA frame.

\begin{figure}[htbp]
\centerline{\includegraphics[width=0.35\textwidth]{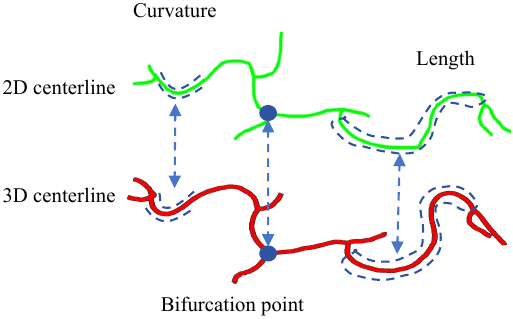}}
\caption{Illustration of anatomical features used for correspondence establishment used in SA-PnP, including bifurcation points, local curvature, and branch lengths. These features are extracted from 2D and 3D centerlines and matched to establish robust initial rigid alignment between modalities.}
\label{fig:PnP}
\end{figure}

\subsection{Global Alignment: SA-PnP}

We introduce SA-PnP, a global initialization method that rapidly establishes correspondences between 2D and 3D points using anatomy information. As illustrated in Fig.~\ref{fig:PnP}, we extract bifurcation points from the vascular tree as auxiliary landmarks to guide alignment. Furthermore, to address potential mismatches in branch correspondence between 2D and 3D representations, we automatically establish branch mappings by comparing anatomical features such as length and curvature. Bifurcations are given higher weights in the registration process due to their uniqueness, while endpoints and outliers are given lower weights or excluded from the optimization.

Formally, the optimal camera pose $(R, t)$ is estimated by minimizing the weighted reprojection error:
\begin{equation}
(R, t) = \arg\min_{R, t} \sum_{i=1}^N w_i \left| \pi\big( K [R \mid t] \mathbf{X}_i \big) - \mathbf{x}_i \right|^2,
\end{equation}
where $K$ denotes the camera intrinsic matrix, and $\pi(\cdot)$ is the perspective projection operation. 
% This stage produces robust temporally aligned rigid samples, with each frame containing camera pose parameters, 2D branch annotations, and projected 3D structures to support downstream vessel deformation estimation.
This stage generates robust, temporally aligned rigid samples, where each frame includes camera pose parameters, 2D branch annotations, and projected 3D structures—providing essential support for subsequent vessel deformation estimation.

\subsection{Local Refinement: TempDiffReg}
Following global alignment, we perform branch-wise registration to capture local deformations of the vasculature. At this stage, the camera parameters, including intrinsics ($K$), rotation ($R$), and translation ($t$), are fixed, and only the 3D projection points are allowed to deform. In addition, to ensure consistent point-wise correspondence between the 2D vessel shape and its 3D projection, both the projected 3D points and the associated 2D annotations are uniformly resampled to a fixed number of $N$ points.

\subsubsection{Encoding Block for Each Branch Pair}
For each 2D–3D vessel branch pair (as established through rigid initialization), we construct a temporal input representation over $N$ consecutive frames. This encoding block is designed to capture geometric and contextual features across time, preparing a unified latent representation for downstream shape prediction.

\begin{figure}[htb]
\centerline{\includegraphics[width=0.4\textwidth]{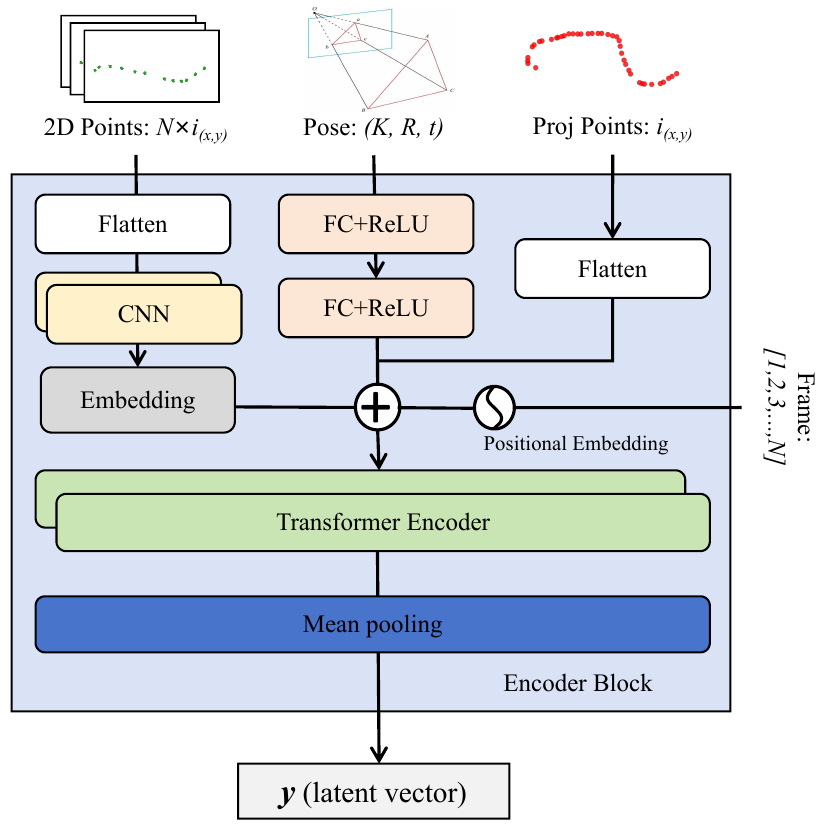}}
\caption{
Architecture of the encoding block for each vessel branch. Inputs include 2D centerlines, projected 3D points, camera pose parameters, and frame indices. These are embedded and fused with positional embedding, then processed by a Transformer encoder to model temporal dependencies. A mean-pooled latent vector $\mathbf{y}$ is extracted and used as a conditioning input to the diffusion-based decoder during shape restoration.
}
\label{fig:model}
\end{figure}

As illustrated in Fig.~\ref{fig:model}, each frame provides four types of input features: the 2D centerline, the projected 3D points, the camera pose parameters (including intrinsics $K$ and extrinsics $R$, $t$), and the frame index. 
Specifically, $t$ 2D points with consecutive $N$ frames are flattened and passed through two convolutional layers, followed by learning branch embedding to enrich spatial features. The pose parameters are encoded using a two-layer fully connected network with ReLU activations. The projected 3D points are also flattened to preserve their spatial structure. Meanwhile, the frame index is converted into a learnable positional embedding to retain temporal ordering. All encoded features from a frame are concatenated and passed through a Transformer encoder (stacked twice) to model temporal dependencies across frames.

Finally, the output tokens from all frames are aggregated via mean pooling to obtain a global latent vector $\mathbf{y}$, which encapsulates the spatiotemporal features of the entire vessel branch sequence. This conditional representation serves as input to the subsequent diffusion-based shape decoder.

\subsubsection{Conditional Diffusion Framework}

\begin{figure*}[htb]%
\centering
\includegraphics[width=0.9\textwidth]{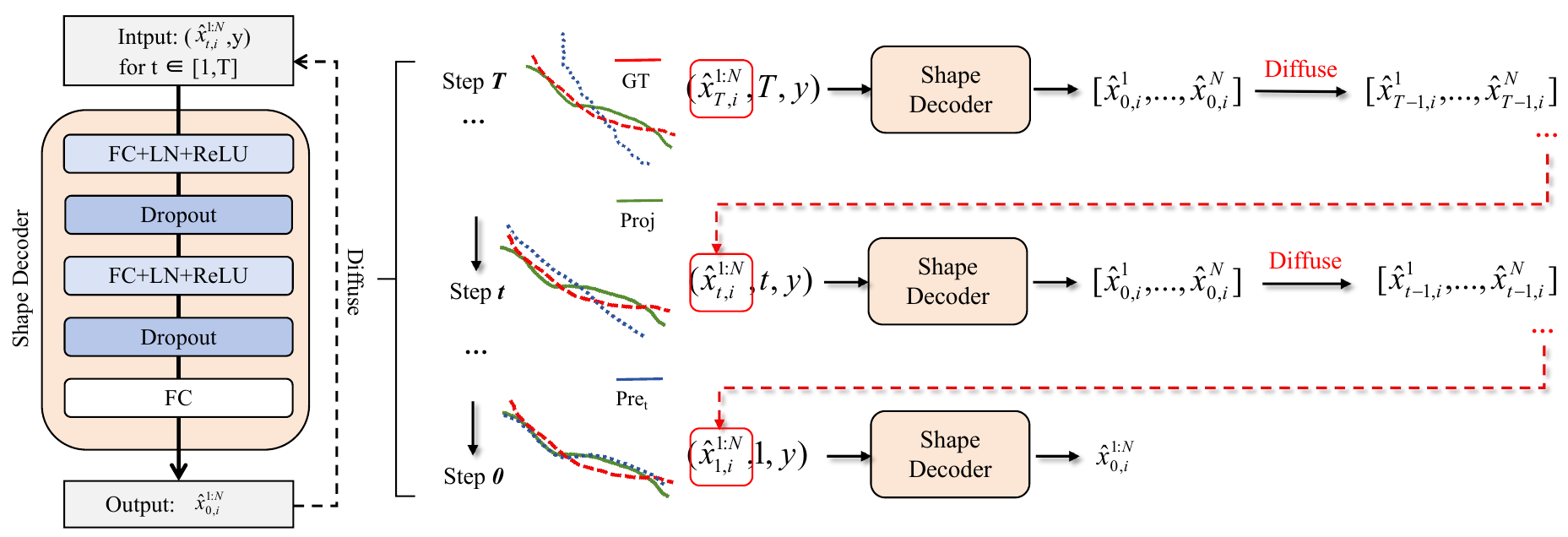}
\caption{Architecture of the conditional diffusion-based shape decoder. At each diffusion step $t \in [1, T]$, the decoder predicts a denoised vessel shape $\hat{\mathbf{x}}^{1:N}_{0,i}$ from the noisy input $\mathbf{x}^{1:N}_{t,i}$, conditioned on the global latent vector $\mathbf{y}$. During training, intermediate predictions are optionally re-noised for supervision across multiple diffusion steps. At inference, the decoder begins from pure Gaussian noise $\mathbf{x}^{1:N}_{T,1}$ and iteratively restores anatomically plausible 2D centerlines. The conditioning vector $\mathbf{y}$ is injected at every time step to ensure temporally consistent and anatomically faithful shape generation.}
\label{fig:decoder}
\end{figure*}

We adopt a conditional denoising diffusion probabilistic model (DDPM)~\cite{ho2020denoising} to learn non-rigid deformations from projected 3D vessel points to target 2D label curves. The goal is to iteratively refine a noisy version of the 2D vessel shape into an anatomically plausible target, conditioned on latent features derived from 3D projections and associated camera pose.

The diffusion process consists of two stages: a forward process that gradually adds Gaussian noise to the target shape, and a reverse process that learns to denoise and recover the shape conditioned on structural priors.

\vspace{0.5em}
\noindent\textbf{Forward Process.}
Let $\mathbf{x}_{0,i}^{1:N}$ denote the ground-truth 2D coordinates of the $i$-th point across $N$ consecutive frames. Let $\bar{\alpha}t = \prod{s=1}^t \alpha_s$ represent the cumulative product of the noise schedule. At each diffusion step $t \in {1, \dots, T}$, Gaussian noise is gradually added to the original vessel shape, resulting in the following forward process~\cite{ho2020denoising}:

\begin{align}
    q(\mathbf{x}_{t,i}^{1:N} \mid \mathbf{x}_{0,i}^{1:N}) = \mathcal{N}(\mathbf{x}_{t,i}^{1:N}; \sqrt{\bar{\alpha}_t} \, \mathbf{x}_{0,i}^{1:N}, (1 - \bar{\alpha}_t) \, \mathbf{I}).
\end{align}

\vspace{0.5em}
\noindent\textbf{Reverse Process.}
The reverse denoising process is modeled as a conditional Markov chain that recovers the original vessel label by progressively removing noise. It is parameterized as:
\begin{align}
    p_\theta(\mathbf{x}_{t-1,i}^{1:N} \mid \mathbf{x}_{t,i}^{1:N}, \mathbf{y}) 
    &= \mathcal{N}\Big(
        \mathbf{x}_{t-1,i}^{1:N}; 
        \notag\\
    &\quad \boldsymbol{\mu}_\theta(\mathbf{x}_{t,i}^{1:N}, \mathbf{y}, t), \;
          \boldsymbol{\Sigma}_\theta(\mathbf{x}_{t,i}^{1:N}, \mathbf{y}, t)
    \Big)
\end{align}

Here, $\mathbf{y}$ is the conditioning vector computed from the encoder, encapsulating spatiotemporal context from multi-frame 3D projections and pose information.

\vspace{0.5em}
\noindent\textbf{Training Objective.}
During training, a diffusion step $t$ is randomly sampled. The noisy input $\mathbf{x}_{t,i}^{1:N}$ is generated via the forward process, and the decoder $f_\theta$ is trained to predict the denoised target $\hat{\mathbf{x}}_{0,i}^{1:N}$ given the condition $\mathbf{y}$:
\begin{align}
    \hat{\mathbf{x}}_{0,i}^{1:N} = f_\theta(\mathbf{x}_{t,i}^{1:N}, t, \mathbf{y}).
\end{align}
To facilitate multi-step supervision, intermediate predictions can be re-noised to obtain $\hat{\mathbf{x}}_{t-1,i}^{1:N}$, allowing training signals to propagate across diffusion steps, as illustrated in Fig.~\ref{fig:decoder}.

\vspace{1em}
\subsubsection{Shape Decoder Architecture}

As illustrated in Fig.~\ref{fig:decoder}, the shape decoder is designed as a lightweight multi-layer perceptron (MLP) that transforms noisy 2D centerlines into anatomically plausible vessel shapes. At each diffusion step $t$, it takes $\mathbf{x}_{t,i}^{1:N}$ as input with condition $\mathbf{y}$ and produces the denoised sample $\hat{\mathbf{x}}_{0,i}^{1:N}$. As analyzed in~\cite{ho2020denoising,ramesh2022hierarchical}, directly predicting the denoised sample at each diffusion step—rather than the noise—has been shown to yield superior performance.

\subsubsection{Training Loss}
% \textcolor{red}{Consider whether to use "$\mathbf{x}^{1:N}_{t,i}$" or not}
The model is optimized with a composite loss that encourages accurate point-wise regression, curve consistency, and noise modeling~\cite{zhuo2024diffusereg,tajdari2022feature}:
\begin{align}
    \mathcal{L}_\mathrm{total} = 
        \lambda_\mathrm{MSE} \mathcal{L}_\mathrm{MSE} +
        \lambda_\mathrm{curv} \mathcal{L}_\mathrm{curv} +
        \lambda_\mathrm{diff} \mathcal{L}_\mathrm{diff}.
\end{align}

Let $i$ denote the number of ordered points in a 2D vessel branch. $\hat{\mathbf{x}}_{0,i}$ and $\hat{\mathbf{x}}_{0,i}$ are the ground truth and predicted coordinates of the $i$-th point at denoising step $t=0$, respectively.
\begin{itemize}
    \item Point-wise MSE:
    \begin{align}
    \mathcal{L}_\mathrm{MSE} = \frac{1}{N} \sum_{i=1}^N \left\| \hat{\mathbf{x}}_{0,i} - \mathbf{x}_{0,i} \right\|^2.
    \end{align}
    This term penalizes Euclidean distance between the predicted and ground truth 2D coordinates.

    \item Curvature Consistency:
    \begin{align}
    \mathcal{L}_\mathrm{curv} = \frac{1}{N-2} \sum_{i=2}^{N-1} &\left| \kappa(\hat{\mathbf{x}}_{0,i-1}, \hat{\mathbf{x}}_{0,i}, \hat{\mathbf{x}}_{0,i+1}) \right. \notag \\
    &\left. -\ \kappa(\mathbf{x}_{0,i-1}, \mathbf{x}_{0,i}, \mathbf{x}_{0, i+1}) \right|,
    \end{align}
    and curvature is computed as:
    \begin{align}
    \kappa(\mathbf{a}, \mathbf{b}, \mathbf{c}) = \frac{2 \cdot |\mathrm{Area}(\mathbf{a}, \mathbf{b}, \mathbf{c})|}{\|\mathbf{a} - \mathbf{b}\| \cdot \|\mathbf{b} - \mathbf{c}\| \cdot \|\mathbf{c} - \mathbf{a}\|},
    \end{align}
    where $\mathrm{Area}(\mathbf{a}, \mathbf{b}, \mathbf{c})$ denotes the area of the triangle formed by the three points $\mathbf{a}$, $\mathbf{b}$, and $\mathbf{c}$ in 2D space, computed using a standard determinant-based formula.
    
    \item Diffusion Prediction:
    \begin{align}
    \mathcal{L}_\mathrm{diff} = \mathbb{E}_{t\in [1, T]} \left[ \left\| \mathbf{\hat{x}}_{0,i}^{1:N} - \mathbf{x}_{0,i}^{1:N} \right\|^2 \right],
    \end{align}
    where $\mathbf{x}_{0,i}^{1:N}$ and $\mathbf{\hat{x}}_{0,i}^{1:N}$ are the target and predicted 2D vessel deformation at diffusion step $t$.
\end{itemize}

\subsubsection{Inference}
During inference, the model starts from a pure Gaussian noise sample $\mathbf{\hat x}_{T,i}^{1:N} \sim \mathcal{N}(0, \mathbf{I})$. Then at the diffusion step $t$, the model denoises $\mathbf{\hat x}_{t,i}^{1:N}$ conditioned on $\mathbf{y}$ to predict the $\hat{\mathbf{x}}_{0,i}^{1:N}$. This output is then further denoised to $\hat{\mathbf{x}}_{t-1,i}^{1:N}$, which serves as the input for the next step $t-1$ in the reverse diffusion process, as shown on the right side of Fig.\ref{fig:decoder}. The stochasticity of the sampling process enables generation of diverse shape hypotheses under the same condition $\mathbf{y}$. This property can be leveraged for uncertainty quantification, which is particularly valuable in ambiguous vascular regions or low-quality image frames.

\begin{figure*}[htb!]%
\centering
\includegraphics[width=1\textwidth]{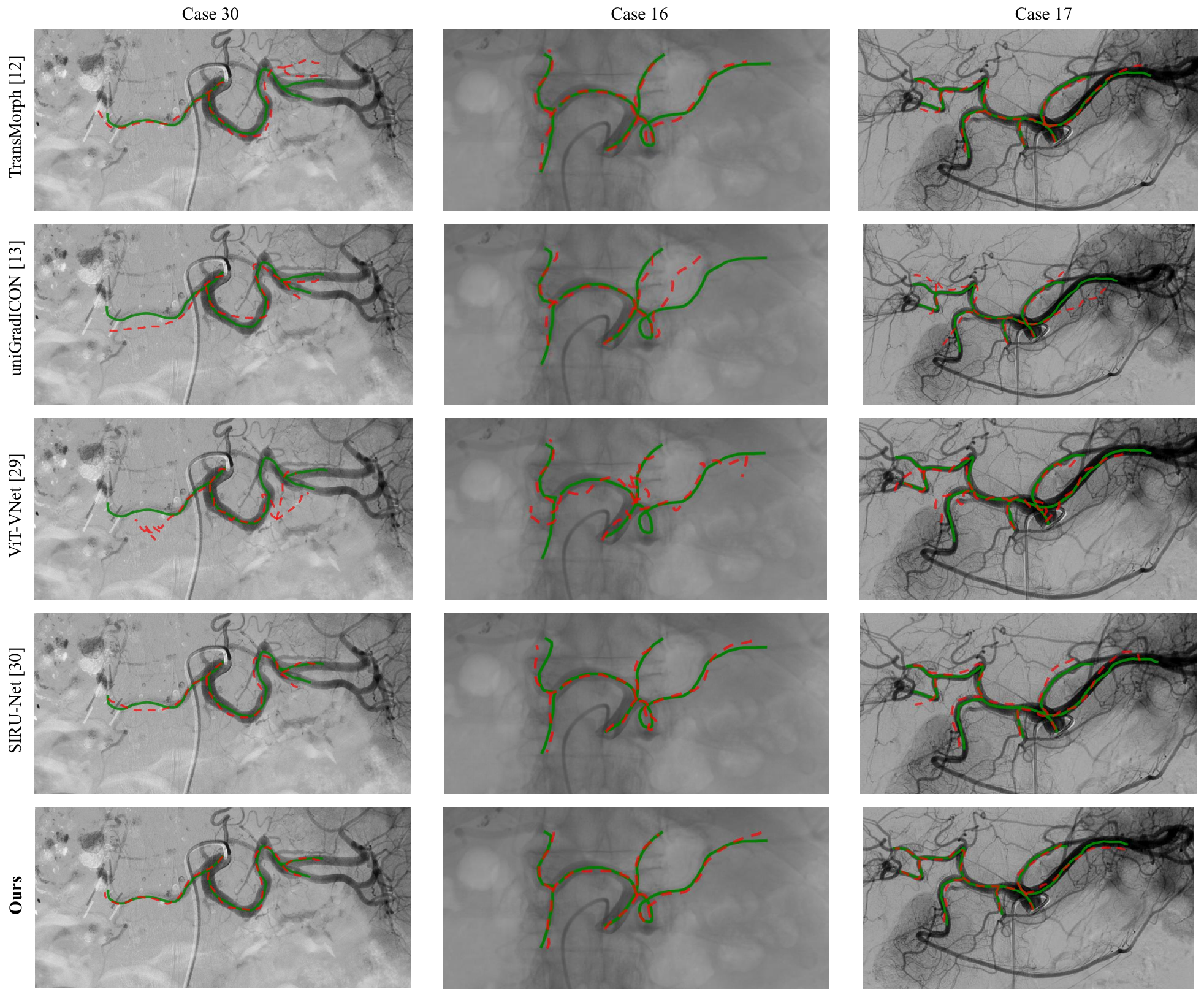}
\caption{
Comparison of registration results on intra-operative 2D DSA images from three representative cases (Case 30, 16, and 17). Ground-truth vessel annotations are shown in green, while registered projections from each method are shown in red dashed lines. Compared methods include TransMorph, uniGradICON, ViT-VNet, SIRU-Net, and the proposed TempDiffReg (bottom row).
}
\label{fig:result}
\end{figure*}

Compared to existing methods~\cite{Rombach_2022_CVPR,zou2025generating}, the proposed framework provides two notable improvements. First, the use of a multi-step diffusion process enables iterative refinement of vessel shape across $T$ denoising steps, rather than relying on a one-shot prediction. This progressive alignment strategy facilitates gradual correction of spatial errors and better handles anatomical ambiguity. As shown in Fig.~\ref{fig:result}, the resulting predictions exhibit smoother trajectories and improved anatomical plausibility.
Second, the inherent stochasticity of the diffusion model supports diverse registration hypotheses at inference time. This is especially beneficial in cases with limited visual information or ambiguous correspondences, where multiple plausible alignments may exist. As illustrated in Fig.~\ref{fig:uncertainty_vis}, the model produces spatially coherent variations that reflect uncertainty in vessel structure, aligning well with the variability observed in clinical practice.

\section{Experiments and Results}

\subsection{Dataset Construction}
We curated a single-center dedicated liver vessel registration dataset, collected from 23 cases of hepatocellular carcinoma (HCC), each containing intra-operative DSA sequences and corresponding pre-operative 3D reconstructions. For each case, only those vessel branches and DSA frames where the anatomy is clearly visible and consistently annotated were retained for model development.

\textbf{Sequential Multi-frame Dataset:}  
we group annotations from each branch of 4 frames to form one multi-frame sample, allowing temporal supervision. Each sample also includes the corresponding camera parameters and projected 3D points. Each clinical case thus provides multiple sequential examples, substantially increasing data diversity and supporting robust model evaluation. In total, we curated 626 such multi-frame samples to train TempDiffReg.

\textbf{Single-frame Dataset for Comparison:}  
For a fair comparison with other methods, which do not leverage temporal information, we also constructed a single-frame dataset. Each sample consists of the 2D vessel annotation, the projected 3D points, and the pose parameters, yielding a total of 1322 samples. This ensures that all baseline methods are trained and evaluated under consistent clinical conditions.

\begin{table}[htbp]
    \centering
    \caption{Comparison with SOTA methods on the single-frame vessel registration dataset (unit: mm). Lower is better.}
    \label{tab:sota_comparison}
    \footnotesize
    \setlength{\tabcolsep}{1mm} 
    \begin{tabular}{|l|c|c|c|c|c|}
       \hline
    Method & MSE $\downarrow$ & MAE $\downarrow$ & MaxErr $\downarrow$ & LenErr $\downarrow$ & CurvErr $\downarrow$ \\
       \hline
    TransMorph~\cite{chen2022transmorph}  & 2.25 & 0.88 & 3.14 & 1.75 & 0.303 \\
    uniGradICON~\cite{tian2024unigradicon} & 1.87 & 0.64 & 3.19 & 2.22 & 0.239 \\
    ViT-VNet~\cite{vitvnet}                & 5.55 & 1.17 & 5.14 & 17.21 & 0.116 \\
    SIRU-Net~\cite{SIRU}                   & 1.89 & 0.62 & 2.58 & 1.47 & 0.124 \\
    \textbf{Ours}                          & \textbf{0.63} & \textbf{0.51} & \textbf{2.01} & \textbf{1.15} & \textbf{0.095} \\
     \hline
    \end{tabular}
\end{table}

\begin{table*}[htbp]
    \centering
    \caption{Extended ablation study: single and two-module variants. Lower values indicate better performance.}
    \label{tab:ablation}
    \footnotesize
    \setlength{\tabcolsep}{2.8mm}
    \resizebox{0.96\textwidth}{!}{
    \begin{tabular}{|l|c|c|c|c|c|}
    \hline
    Variant & MSE (mm) $\downarrow$ & MAE (mm) $\downarrow$ & MaxErr (mm) $\downarrow$ & LenErr (mm) $\downarrow$ & CurvErr $\downarrow$ \\
    \hline
    w/o Temporal Modeling                  & 1.12 & 0.77 & 2.89 & 1.653 & 0.143 \\
    w/o Structural Prior                   & 1.06 & 0.74 & 2.71 & 1.632 & 0.145 \\
    w/o Transformer Encoder                & 0.91 & 0.68 & 2.45 & 1.410 & 0.121 \\
    w/o Diversity Loss                     & 0.75 & 0.56 & 2.33 & 1.192 & 0.104 \\
    \hline
    w/o Temporal + Transformer             & 1.39 & 0.88 & 3.11 & 1.799 & 0.162 \\
    w/o Temporal + Structural Prior        & 1.33 & 0.85 & 2.98 & 1.765 & 0.158 \\
    w/o Structural Prior + Diversity       & 1.19 & 0.79 & 2.81 & 1.701 & 0.150 \\
    w/o Transformer + Diversity            & 1.04 & 0.72 & 2.56 & 1.480 & 0.125 \\
    \hline
    \textbf{Full Model (Ours)}             & \textbf{0.63} & \textbf{0.51} & \textbf{2.01} & \textbf{1.145} & \textbf{0.095} \\
    \hline
    \end{tabular}}
\end{table*}

\subsection{Evaluation Metrics}
We use mean squared error (MSE), mean absolute error (MAE), and maximum error (MaxErr) to quantify the geometric accuracy of predicted vessel curves relative to ground truth annotations, capturing both average and worst-case deviations at the point level.

To evaluate structural fidelity, we report length error (LenErr), which measures discrepancies in the overall arc length between predicted and reference centerlines, and curvature error (CurvErr), which captures local geometric differences by comparing curvature profiles along the curves~\cite{chen2025survey}.

\subsection{Comparison with SOTA Methods}
To quantitatively assess the proposed method, we compare it against SOTA methods. All models are trained and evaluated on the same single-frame dataset comprising 1322 samples, using identical data splits, normalization procedures, and evaluation protocols to ensure a fair and consistent comparison. The following methods are selected as baselines: TransMorph~\cite{chen2022transmorph}, uniGradICON~\cite{tian2024unigradicon}, ViT-VNet~\cite{vitvnet}, and SIRU-Net~\cite{SIRU}. For each baseline, we adapt the official implementation to handle vessel centerline data and follow the recommended hyperparameter settings.

Our proposed method achieves superior performance across all evaluation metrics compared to SOTA methods. Specifically, it attains the lowest MSE (0.63 mm), and MAE (0.51 mm), indicating high point-wise accuracy in aligning 2D vessel labels with their 3D counterparts. Furthermore, our model yields the smallest maximum error (2.01 mm), reflecting robust behavior even in challenging cases involving large deformations or imaging artifacts.

In terms of structural fidelity, the proposed method achieves the lowest LenErr (1.145 mm) and CurvErr (0.095), demonstrating its ability to preserve both global vessel topology and fine-grained geometric details. Notably, existing methods like TransMorph and SIRU-Net achieve relatively strong results on some metrics, but they are fundamentally limited by their reliance on single-frame input and the lack of explicit curvature modeling. Our temporal diffusion framework, by contrast, benefits from multi-frame context and shape-aware supervision, resulting in consistently better accuracy and anatomical plausibility across diverse registration scenarios.

As shown in Fig.~\ref{fig:result}, our method consistently achieves the most accurate and anatomically plausible registration results. Compared with TransMorph and SIRU-Net, our method produces smoother trajectories with better alignment, particularly around vascular bifurcations. Although TransMorph and SIRU-Net also exhibit relatively small alignment errors, they show minor deviations near branching regions. In contrast, uniGradICON introduces notable distortions in curve shape, leading to anatomically implausible warping in all cases. ViT-VNet suffers from curve discontinuity and topological inconsistencies, especially visible in Case 30 and Case 16.

\subsection{Ablation Study}

To further understand the contribution of each core component in our temporal diffusion-based vessel registration framework, we conducted a series of ablation experiments. Specifically, we systematically removed single- and dual-module to assess their impact on registration performance. All ablation variants were trained and evaluated using the same dataset and metrics as the full model.
\begin{itemize}
    \item \textbf{Temporal Modeling}: Replaces multi-frame input with single-frame branch input, removing explicit temporal context.
    \item \textbf{Structural Prior}: Disables both length and curvature losses, relying solely on point-wise MSE supervision.
    \item \textbf{Transformer Encoder}: Replaces the temporal Transformer module with a simple MLP-based decoder.
    \item \textbf{Diversity Loss}: Removes the noise diversity term $\mathcal{L}_\mathrm{diff}$ during training.
\end{itemize}

The results in Table~\ref{tab:ablation} highlight the importance of each component. Removing temporal modeling leads to the most significant performance drop across all metrics, confirming the value of multi-frame input for capturing motion continuity and anatomical consistency. Disabling structural priors (i.e., length and curvature losses) also results in notable degradation, especially in geometric consistency metrics such as CurvErr and LenErr. The removal of the Transformer encoder impairs the model’s ability to model long-range temporal or spatial dependencies, increasing maximum and structural errors. Furthermore, removing the diversity loss slightly reduces performance, particularly in maximum error, reflecting a decreased ability to explore and refine complex deformation hypotheses during training.

Among ablations of dual-module, removing both temporal modeling and the Transformer leads to the worst results (MSE: 1.39 mm), confirming their complementary roles in spatiotemporal modeling. Other combinations also cause noticeable degradation, reinforcing the necessity and effectiveness of each design choice. Overall, the full model consistently outperforms all ablation variants, demonstrating that the combination of temporal context, anatomical awareness, and shape-aware regularization is critical for achieving robust and precise vessel registration.

\begin{figure}[htbp]
    \centering
    \includegraphics[width=0.80\linewidth]{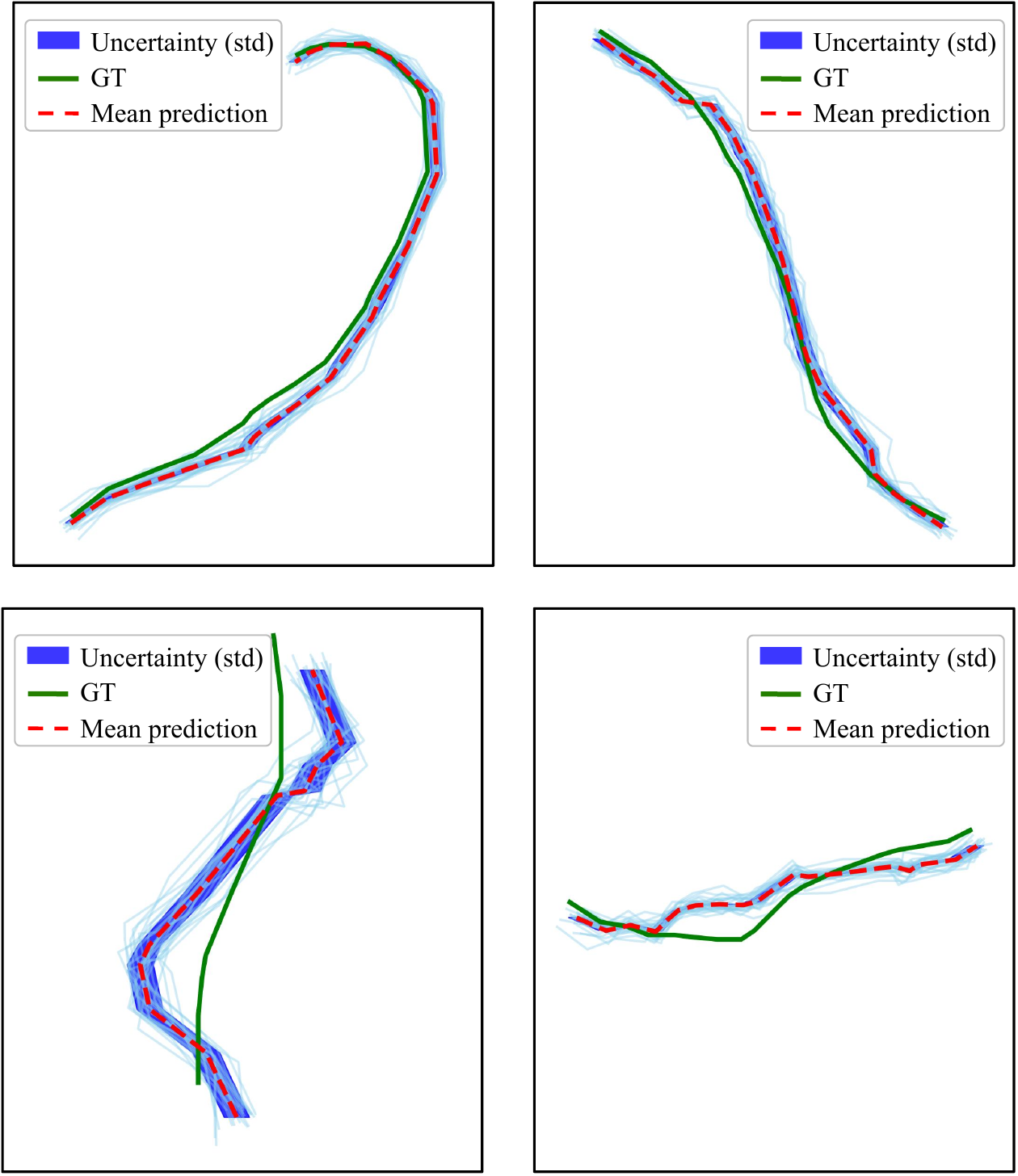}
    \caption{Prediction uncertainty visualization for a representative vessel branch. The green curve represents the ground truth (GT), the red dashed line is the mean prediction, and the light blue curves show individual stochastic samples. The shaded blue region indicates pointwise standard deviation, highlighting areas of high anatomical ambiguity or model uncertainty, such as bifurcations or occlusions.
    }
    \label{fig:uncertainty_vis}
\end{figure}

\subsection{Uncertainty and Diversity Evaluation}

To further analyze the expressiveness and reliability of our diffusion-based vessel registration model, we assess its output diversity and structural uncertainty through stochastic sampling. For each vessel branch in the test set, we perform $N=25$ independent samplings under identical input conditions, yielding a set of predicted curves. As shown in Fig.~\ref{fig:uncertainty_vis}, the results showed that proposed method produces highly consistent predictions in anatomically well-constrained regions, with narrow uncertainty bands indicating high confidence. In contrast, broader uncertainty appears in regions with ambiguous structures, such as bifurcations, occluded areas, or high-curvature segments. This pattern demonstrates the model’s ability to capture localized uncertainty and adapt to varying anatomical complexity. The stochastic generation mechanism thus not only enables structural diversity but also provides an interpretable confidence measure for downstream clinical applications.

\section{Discussion}

Experimental comparisons demonstrate that our proposed method consistently outperforms SOTA approaches across all five quantitative metrics, reducing MSE by 66.7\% (vs. SIRU-Net), MAE by 17.7\% (vs. SIRU-Net), MaxErr by 22.1\% (vs. SIRU-Net), LenErr by 22.0\% (vs. SIRU-Net), and CurvErr by 18.1\% (vs. ViT-VNet). As visualized in Fig.~\ref{fig:result}, TempDiffReg produces smoother and more anatomically consistent deformations, particularly around bifurcations and high-curvature regions, where competing methods often exhibit local distortions or topological mismatches. These results validate the effectiveness of integrating temporal modeling and branch-wise deformation, especially in capturing complex vascular motion that is challenging to resolve from single-frame inputs.

The ablation study provides deeper insights into the contribution of each component. Among all variants, removing temporal modeling leads to the largest performance drop, emphasizing the critical role of multi-frame context in capturing motion continuity and resolving inter-frame ambiguities. For dual-module ablations, removing both temporal modeling and the Transformer leads to the worst results, and other combinations also cause noticeable degradation, underscoring the necessity and effectiveness of full model.

Additionally, as shown in Fig.~\ref{fig:uncertainty_vis}, repeated sampling under identical inputs produces a family of plausible vessel shapes, with point-wise uncertainty bands highlighting regions of structural ambiguity. This diversity-aware behavior not only enables uncertainty quantification but also enhances model interpretability, providing clinicians with richer information.

Despite promising results, two key limitations remain. First, the dataset size is limited due to the difficulty of acquiring paired DSA and CTA. Secondly, metal induced artifacts caused by stents or clips will reduce image quality and affect registration accuracy. To address these issues, future work will focus on expanding the dataset through additional clinical collection and simulation-based data augmentation~\cite{ou2022deep,liu2024unsupervised}. Moreover, integrating artifact-robust vessel segmentation algorithms could enhance the method's applicability in real-world clinical scenarios with severe image degradation.

\section{Conclusion}
In this work, we presented TempDiffReg, a temporal diffusion-based framework for 2D–3D vessel registration, designed to anatomically model accurate deformations across sequential angiographic frames. 
TempDiffReg is the first data-driven method to unify three critical components for this task, including temporal conditioning, anatomical branch modeling, and structured shape diffusion. By incorporating multi-frame sequences, the model captures motion continuity, resolves ambiguities present in single-frame observations, and preserves vascular topology across time.
By integrating anatomical priors into a coarse initialization and leveraging temporal continuity for fine-grained refinement, TempDiffReg outperforms all baseline methods across five evaluation metrics on a private clinical dataset, confirming its effectiveness in achieving precise and anatomically consistent 2D–3D registration.
%summary
These findings highlight the effectiveness of combining temporal modeling with anatomical priors in diffusion-based registration, offering a promising solution for enhancing navigation and guidance in interventional procedures.

\bibliographystyle{unsrt}
\bibliography{Regrefs}

\end{document}